# Edge-GPU Based Face Tracking for Face Detection and Recognition Acceleration

Asma Baobaid and Mahmoud Meribout, Senior *Member, IEEE*

*Abstract*— Cost-effective machine vision systems dedicated to real-time and accurate face detection and recognition in public places are crucial for many modern applications. However, despite their high performance, which could be reached using specialized edge or cloud AI hardware accelerators, there is still room for improvement in throughput and power consumption. This paper aims to suggest a combined hardware-software approach that optimizes face detection and recognition systems on one of the latest edge GPUs, namely NVIDIA Jetson AGX Orin. First, it leverages the simultaneous usage of all its hardware engines to improve processing time. This offers an improvement over previous works where these tasks were mainly allocated automatically and exclusively to the CPU or, to a higher extent, to the GPU core. Additionally, the paper suggests integrating a face tracker module to avoid redundantly running the face recognition algorithm for every frame but only when a new face appears in the scene. The results of extended experiments suggest that simultaneous usage of all the hardware engines that are available in the Orin GPU and tracker integration into the pipeline yield an impressive throughput of 290 FPS (frames per second) on 1920 x 1080 input size frames containing in average of 6 faces/frame. Additionally, a substantial saving of power consumption of around 800 mW was achieved when compared to running the task on the CPU/GPU engines only and without integrating a tracker into the Orin GPU's pipeline. This hardware-codesign approach can pave the way to design high-performance machine vision systems at the edge, critically needed in video monitoring in public places where several nearby cameras are usually deployed for a same scene.

## I. INTRODUCTION

VIDEO analytics systems are recently deployed in various domains including healthcare, industries and public places where it is used for the detection and recognition of diseases, objects and people. Among the different uses of video analytics, face recognition remains one of the critical and commonly used application due to its ability to identify and recognize faces for numerous applications. Face recognition can be deployed in small areas like companies for attendance systems to large public areas and critical infrastructures like malls, parks airports and government buildings for security and surveillance reasons. Existing face recognition systems are mainly based on algorithms such as local binary patterns (LBP) [1], principal component analysis

(PCA) [2], support vector machines (SVM) [3], and Gabor Filter [4]. These algorithms have been successfully implemented in [5], [6], [7] and were capable of achieving around 10 FPS on cloud GPU devices. Improvements in processing time were further achieved with the recent development of deep learning algorithms like Convolutional Neural Networks (CNNs) based models that are designed to run on different advanced hardware accelerators mainly GPUs and FPGAs. However, due to built-in memory limitation in FPGAs which limit the device to only host light and less accurate CNN-based models, GPUs are usually more preferred in applications were complex CNN models are required [2]. Despite the huge improvements in CNN-based models and associated computational hardware engines, real-time face recognition systems are still challenging to design especially for crowded, large-scale areas where multiple video streams and large number of people exist. One effective way that was considered in several studies to control large scale areas faces monitoring is through adopting the face tracker to improve the face detection systems [8][9][10]. However, this approach was not considered for face recognition systems to track recognized faces and avoid repetitive recognition. Another advantage of face tracking is that usually, in a streaming video, a recurrence of faces exists in several frames; if this system is used for video surveillance where an immediate alarm must be raised with each security threat, an alarm will be sent to the end user each time a face is identified (i.e., each frame) which causes duplicated uncontrolled alarms. Integrating a tracker in such systems will improve performance and minimize duplicated alarms by sending a single alarm per track [8]. Additionally, most face recognition systems are not capable to recognize faces with non-frontal face images, thus, integrating a tracker into such a system will enhance recognition by linking faces that gradually change from frontal view to profile view where it is challenging for face recognition models to recognize [11]. The other room for improvement, considered in this paper, is maximizing the simultaneous usage of all the hardware engines available nowadays in edge GPUs. This enhancement of the level of parallelism yields higher performance than other similar works that use exclusively GPU or CPU cores [5][6][7]. In summary, the contributions of this paper can be described as follows:

1. Almost all existing works of face recognition systems using GPU hardware accelerators were designed to be executed only on the GPU and CPU engines of the device

Asma Baobaid and Mahmoud Meribout are with the Electrical & Computer Engineering Department, Khalifa University, Abu Dhabi, UAE (email: asma.baobaid@outlook.com and mahmoud.meribout@ku.ac.ae)

while ignoring other hardware engines available in heterogeneous embedded GPU devices [12][13][14]. This paper targets the utilization of other hardware engines that are available in Jetson devices, including Deep Learning Accelerators (DLAs), Vision Image Compensator (VIC), Video Decoder/Encoder (NVDEC/NVENC), CPUs. In addition to the GPU CUDA and Tensor Cores. The paper also leverages weights quantization and hardware allocation techniques to maximize system performance.

2. A tracking algorithm between the detection and recognition stages is added to reduce both the processing time and the power consumption without altering the system accuracy. This approach was not considered in other previous works. Three trackers were assessed regarding load consumption and robustness, with the optimal one selected to avoid repetitive identification, improving pipeline throughput.

3. Previous research works have used available datasets to train and test the developed models where these datasets consist of frontal face images. However, in a real-time face recognition system targeting public places, the system must detect and recognize faces with different orientations and scale. This paper addresses this constraint and builds and assesses a dataset that includes a traditional dataset from the Middle East region, which was not considered in earlier related research works.

## II. BACKGROUND

Face recognition systems mainly consist of three main stages, detection, feature extraction and face recognition. The face detection stage generates bounding boxes around faces in each frame. Face detection algorithms like Viola-Jones, Haar-Cascade Classifier or CNN-based like Multitask Cascaded Convolutional Networks (MTCNN) [15] were intensively used in the literature. The bounding boxes region highlighted by the detection stage will be sent for the feature extraction stage. The feature extraction stage will work to extract the main features of the face through algorithms like LBP, PCA, Independent Component Analysis (ICA) [16], Linear Discriminant Analysis (LDA) [17] or CNN-based algorithms like FaceNet which surpassed all others in terms of accuracy on LFW dataset [18]. Face features can be represented by binary patterns, eigenvectors or embeddings, which will then require a classification stage to compare the output of feature extraction stage with existing features in the database for either face identification or verification. Lastly, a face can be classified using algorithms like Support Vector Machine (SVM) or k-Nearest Neighbors (KNN). Among all aforementioned algorithms, CNN based proved to be more robust in terms of accuracy [19]. However, one of the main drawbacks is that they are more computationally complex than convolutional models that depend on simple mathematical operations. Nevertheless, their intrinsic parallelizable computation model has motivated researchers to suggest various parallel hardware accelerators, usually GPU- or FPGA-based. For example, in [20], cascaded face detection and recognition system was built based on MTCNN for detection and FaceNet for recognition. The system was tested on 4 cloud-based and 4 edge-based devices by NVIDIA. The lowest processing time achieved using cloud based device is 0.05 seconds per frame (i.e., 20 FPS), where the lowest achieved based on an edge device is 0.27 seconds per frame (i.e., 4 FPS), these were achieved using RTX 2080 Ti and Jetson Xavier AGX, respectively. These results show improvements compared to [21], where the system achieved 1 FPS on the same face detection and face recognition models using Raspberry Pi 3B+. This highlighted the need to run computationally intensive deep learning models on dedicated hardware accelerators like GPUs and FPGAs that can handle parallel computation and provide high-throughput capabilities. Another approach considered by researches to address the issue of the high processing time, is the development of lightweight, less complex neural network models. This includes the FaceDetect model developed by NVIDIA for face detection, which is based on ResNet-18 and can be optimized for edge devices [22] and the two-stage FaceBoxes model developed for face detection through improving the Faster R-CNN model [23]. In [24] FaceBoxes algorithm which is a less intensive model compared to MTCNN, in addition to FaceNet for recognition are used. The system was tested on Jetson Nano and Jetson TX2 where it achieved average of 4 FPS and 7.5 FPS, respectively, on an input video stream comprising a maximum of 3 faces/frame; corresponding to double the throughput achieved in [20]. The reason of this is that the usage of FaceBoxes model resulted in 1.5 times reduction in detection processing time compared to MTCNN model. However, real-time performance, corresponding to a minimum throughput of 30 FPS, still couldn't be achieved in both works. This is due to the underutilization of the NVIDIA GPUs hardware, where only the CPU and GPU cores were used, ignoring other powerful hardware accelerators such as the DLA and VIC engines. Additionally, these papers have neglected the integration of a face tracker into the pipeline and did not consider the fact that almost all faces in a specific frame will continue to appear in the several next frames. Face tracking is mainly applied through the use of object tracking approaches, targeting a face as a region of interest. Some papers have adopted the tracking-by-detection approach; however, these papers only considered face tracking as an improvement to face detection systems without considering the face recognition stage. To illustrate, a multi-face tracker based on CNN model is proposed in [10]. As the developed tracker is based on CNN and feature extraction, the implementation of such models can introduce significant computational complexity to any system which therefore requires a dedicated hardware accelerator. However, the authors did not suggest a hardware accelerator for the system to achieve the real-time performance. Another implementation of tracker-by-detection is proposed in [25]. The author suggested using a Kalman filter to estimate the location of the faces based on an estimator that learns from the previous state. The Kalman filter is based on simpler mathematical modeling than CNN-based trackers and thus requires few hardware resources. The authors used the cloud NVIDIA 16660Ti GPU engine to run the detection and the tracker for testing; however, only a single face per frame is used for testing, where testing multiple faces per frame was suggested as a future recommendation by the authors. Another approach that was developed for object tracking is the use of Discriminative Correlation Filter (DCF) based trackers [26] [27] [28] the principle of which is to track the representative features of an object such as its corners or curvatures of its edges. An improvement to the simple DCF tracker was then suggested in [29] where the authors proposed using multi-channel filters instead of linear correlation filters to enhance the existing pre-developed model. The suggested tracker achieved a mean precision of 72.8% in 50 video sequences and was capable of achieving real-time performance with 292 FPS. A further improvement to the DCF tracker was proposed in [30], where the reliability was enhanced through the integration of the Channel and Spatial Reliability (CSR) assessment. The algorithm developed for enhanced tracker (CSR-DCF) is more computationally complex but yields higher robustness. This is because it takes into consideration the weights of the important features and spatial region of the target, which can then provide a more accurate prediction of the object in the upcoming frame. Another study [31] proposed an alternative improvement in the DCF tracker by using the lasso regression instead of ridge regression in addition to low-rank constraints, which therefore resulted in a simplified model that focuses on the most important features about the tracked objects to improve robustness. Nevertheless, none of the aforementioned systems have considered using the DCF tracker for face tracking, particularly for enhancing the throughput of the face detection and tracking tasks.

## III. Methodology

### A. GPU Hardware Platform

NVIDIA company offers a series of GPUs, the performance of which reaches a few hundred TOPS (tera operations per second). Among the released edge devices, the latest Jetson AGX Orin offers several enhanced hardware features compared to the previous ones (Fig. 1). The processor comprises 16 Streaming Multiprocessor (SMs) with 128-CUDA ampere cores and 64 Tensor Cores. The tensor cores are designed to perform hardwired matrix multiplications involving half-precision (FP16) or 8 bits-integer operations (INT8). The large number of SMs allows a high fine-grained parallelism and can potentially accelerate tasks split into several independent subtasks. In addition, the GPU offers several memory hierarchies where thread-local registers are the fastest, followed by a 192 kB L1 memory cache for each SM block. This memory space is enough to store one video frame. The slowest 4 MB L2 memory is shared by all SMs and is enough to store several video frames which is needed to accelerate the tracking algorithm. The device also comprises 3 CPU clusters and a 4 MB system cache shared among all the 3 clusters. Each CPU Cluster has 4 cores and 2MB L3 cache, whereas each core includes 64 kB instruction L1 cache, 64 kB data cache, and 256 kB of L2 cache. The dual Deep Learning Accelerator (DLA) hardware engine that implements most CNN layers is more power-effective than the GPU engine. Despite performing slightly less than the SMs, its main advantage is to yield a performance/W 2.5 times higher than the SMs by computing a total of 2 x 52.5 INT8 sparse TOPs. The processor also includes hardwired video encoder (NVENC) and decoder (NVDEC) modules; in addition to Orin System-on-Chip, a Video Imaging Compositor (VIC), which supports some low-level image processing tasks such as filtering. It also comprises a Programmable Vision Accelerator (PVA), based on VLIW architecture to efficiently implement convolutions-based image processing tasks such as filtering and stereo vision.

The off-chip 256-bit data bus-64 GB LPDRAM5 memory has the longest latency and can store several frames, which SMs, DLAs, PVA, video encoders, and video decoder engines can use. This memory can be accessed at 204.08 GB/s bandwidth, which is high enough for multi-batch processing since it far exceeds the bandwidth of a single standard video stream (i.e., 30 FPS).

### B. Parallel Hardware Algorithm

Fig.2 shows the block diagram of the suggested face recognition system. It consists of 4 pipeline stages that utilize different hardware engines available on edge Orin GPU hardware. Previous GPU-based face detection and recognition systems did not consider this simultaneous usage of hardware accelerators within the device. The pipeline shows the designed system, from capturing the input video to displaying recognized faces. Currently, most cameras used in public places typically feature an Ethernet interface to generate the output video stream per some international standards, such as H264 or H265. Thus, performing face detection and recognition at the edge requires a dedicated real-time video decoding engine featured in most recent NVIDIA GPUs, including the ORIN GPU. Accordingly, once a frame is captured, it will be sent to the NVDEC hardware engine, which results in very low latency and low power consumption. FaceDetect [22] and FaceNet [15] models are used for detection and recognition, respectively. The face tracker is integrated between the detection and the recognition stages. It is configured to run on the VIC hardware engine of the device, while the GPU and DLA engines are dedicated to run the face detection and recognition deep learning models. This heterogeneous pipeline architecture allows configuration and maximizing hardware utilization. First, each input video stream passes through the detection module, where a unique ID will be assigned to each detected face. The tracker will track the detected faces where only the newly detected faces with new IDs will be passed to the recognition stage for identification. However, if a face was previously detected and identified in previous frames, the same previously identified identity will be assigned to the face. This will prevent the system from re-applying recognition on all faces in every frame. This approach improves previous research works where face recognition is repeatedly applied to all detected faces for each frame. Thus, a reduction in pipeline processing is expected with the tracker's integration. Nevertheless, the allocation of the face detection and recognition CNN-based models into the DLA or the GPU/Tensor cores depends on the structure of the model, as DLA is incapable of hosting all deep learning model layers. The DLA supports convolutional, deconvolutional, fully connected, activation, pooling, and batch normalization layers. However, there are special requirements for these operations to be supported [32]. For any unsupported layer that falls back to the GPU, after execution in the GPU output of this layer has to be transferred to memory to be used by the DLA. Additionally, it is worth mentioning that DLA and Tensor cores only support half-precision (FP16) or integer operations (INT8). Thus, to maximize the usage of the hardware and to enable the DLA and tensor cores engines, the pruned version of the Face Detect model is used with INT8 precision. In contrast, pruning and quantization were applied to the FaceNet model to reduce the precision from FP32 to FP16. The expected time reduction from the hardware allocation strategy is shown in Fig.3 (a) and (b). Fig.3 (b) shows the expected reduction in processing time when running the deep learning models on DLA with support of the GPU, compared to running both models fully on the GPU, as shown in Fig.3 (a). Furthermore, the expected additional time reduction by integrating a tracker is shown in Fig.3 (c).

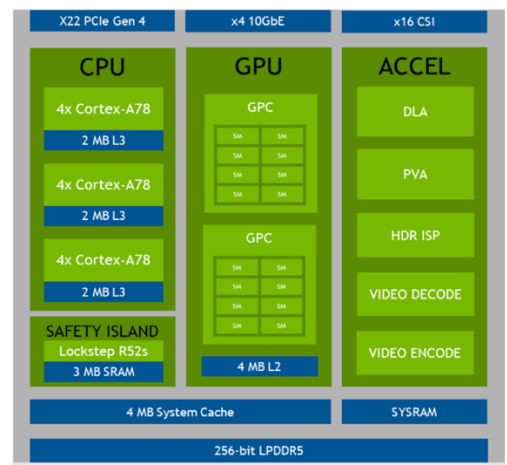

*Figure 1: Orin System-on-Chip (SoC) Block Diagram*

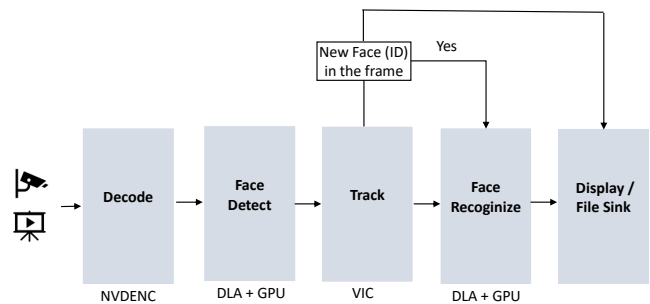

*Figure 2: Face Recognition Model Pipeline*

## C. HARDWARE PARTITIONING STRATEGIES

FaceDetect is a pre-trained model that uses the NVIDIA object detection model, DetectNet_v2, which uses ResNet18 as a backbone for feature extraction. DetectNet_v2 generates bounding boxes on the input image by dividing the input image into a 16x16 grid and then proceeds by generating two tensors, the converge "cov" and bbox. Cov gives the number of cells an object covers, while the bbox tensor determines the four normalized parameters, $x_1$, $y_1$, $x_2$, and $y_2$ of the bounding-box. The base model consists of 18 layers corresponding to convolutional layers, max and average pooling layers, and a fully connected layer, as shown in Fig.4 (a). This structure

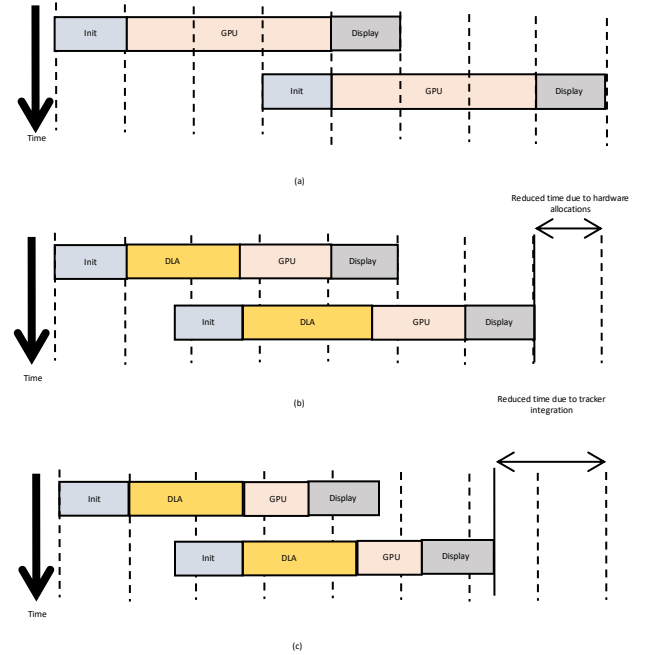

Figure 3: Example of hardware allocation (a) without using pipelining and tracker (b) using the DLA + GPU cores in the pipelining (c) using pipelining and tracker (b)

makes the model suitable for full implementation in both DLA or GPU. The model architecture of FaceNet is presented in Fig. 4 (b). It was originally developed by Google to predict the faces' identities [18]. It extracts high-quality features from the face and predicts a 128-elements vector representation called face embedding. Face embeddings are then mapped to generate a compact Euclidean space, where L2 distances are calculated to measure face similarity [21]. Similarly to FaceDetect, its backbone network consists of a cascade of convolutional, pooling and fully connected layers.

| Type | Output Size |
| --- | --- |
| Conv1 (7x7, stride 2) | 112x112x64 |
| Max Pool (3x3, stride 2) | 56x56x64 |
| Residual Block (2x) | 56x56x64 |
| Residual Block (2x) | 28x28x128 |
| Residual Block (2x) | 14x14x256 |
| Residual Block (2x) | 7x7x512 |
| Avg Pool (7x7) | 1x1x512 |
| Fully Connected | 1x1x1000 |
| Total | |

(a)

| Type | Output Size |
| --- | --- |
| conv1 (7x7×3, 2) | 112x112x64 |
| max pool + norm | 56x56x64 |
| inception (2) | 56x56x192 |
| norm + max pool | 28x28x192 |
| inception (3a) | 28x28x256 |
| inception (3b) | 28x28x320 |
| inception (3c) | 14x14x640 |
| inception (4a) | 14x14x640 |
| inception (4b) | 14x14x640 |
| inception (4c) | 14x14x640 |
| inception (4d) | 14x14x640 |
| inception (4e) | 7x7x1024 |
| inception (5a) | 7x7x1024 |
| inception (5b) | 7x7x1024 |
| avg pool | 1x1x1024 |
| fully conn | 1x1x128 |
| L2 normalization | 1x1x128 |
| Total | |

(b)

Figure 4: Model Structure for (a) FaceDetect-ResNet18 (b) FaceNet

An important design aspect for accelerating the system throughput is the policy to allocate the FaceDetect and FaceNet models onto the GPU and DLA hardware engines. For instance, to select the most efficient hardware allocation, it is important to map supported layers of the models on DLA while minimizing the data transfer between the hardware engines. As was mentioned earlier, DLA supports several layers of CNN models under some conditions. Specifically, in the case of the FaceNet model, compiling the model to run on DLA through TensorRT results in the creation of several shuffle layers. The purpose of a shuffle layer is to transform the format of the output of a given layer to make it compatible with the DLA hardware architecture. Although DLA supports the shuffle layer, generated shuffle layers did not follow the shuffle layers requirements, and therefore, they were allocated to run on GPU. For any unsupported layer that falls back to the GPU, its output must be transferred to the DLA's local memory after being executed in the GPU core. If a non-connected supported layer repeats several times or a model has many separated unsupported layers, the model partitioning will increase device-to-device memory copy (memcpy). As each layer depends on the previous layer results in CNN feed-forward operations, the back-and-forth communication between the DLA and GPU will affect the model processing time

### D. TRACKER INTEGRATION

The face tracker was designed to run on the VIC hardware engine of the device to offload both the GPU and DLAs engines. Three different tracking methods are analyzed, namely the Intersection Over Union (IOU), the Simple Online and Realtime Tracking (SORT), and the Discriminative Correlation Filter (DCF). The IOU is proposed as a lightweight simple tracking algorithm that measures the overlapping of detection boxes between two consecutive frames [33]. The IOU equation can be represented by,

$$IOU\ (a,b) = \frac{Area\ (a) \cap Area(b)}{Area\ (a) \cup Area(b)} \qquad (1)$$

This algorithm assumes that there are no gaps in detection between frames, and it assumes that the bounding boxes of the detected object in two consecutive frames will hugely overlap. Thus, this algorithm is computationally efficient as it uses a simple mathematical equation; however, it is inadequate to track objects exposed to large-scale and rotation changes. Another efficient tracking model is the Simple Online and Realtime Tracking (SORT) model [34] which combines the Kalman Filter and Hungarian algorithm. The SORT model depends on the association between the bounding boxes in two consecutive frames, Kalman filter is then applied to learn from previous frames and predict the next location based on the overlapping and the motion of the detected object. The state prediction of Kalam Filter is mathematically represented by

$$\hat{x}_{k|k-1} = F\hat{x}_{k-1} + \mathbf{B}u_{k-1} \qquad (2)$$

Where F is the state transition matrix, B is the control input matrix, and $u_{k-1}$ is the control vector at iteration $k$-1.
The Kalman filter keeps predicting and correcting the predictions to decrease the uncertainty of its model. The prediction update is mathematically represented by,

$$\hat{x}_{k|k} = \hat{x}_{k|k-1} + \mathbf{K}_k \left( \mathbf{z}_k - \mathbf{H}\hat{x}_{k|k-1} \right) \qquad (3)$$

Where $\mathbf{z}_k$ is the vector measurement (bounding box coordinates), H measurement matrix and $\mathbf{K}_k$ is the Kalman Gain to balance uncertainty between measured and predicted. IOU and SORT algorithms can be executed at high throughput using edge GPUs or even less complex hardware accelerators with enough memory capacity to store the input and output frames. However, they both depend on the IOU overlapping data, which may result in frequent ID-switches [8]. To address this challenge more complex models like the DCF algorithm which was introduced in [29] have represented advancements in terms of robustness. The reason for this, is that DCF-based tracker unlike SORT and IOU, it utilizes the feature representation and correlation filters which depends on target appearance and result in more robust tracking. The correlation filter ($\widehat{CF}$) in Fourier domain is presented as,

$$\widehat{CF} = \frac{\sum \widehat{y_k} \cdot \hat{x}_k^*}{\sum |\hat{x}_k|^2 + \lambda} \qquad (4)$$

Where $\hat{x}_k$ is the Fourier transform of the frame $k$ where the target object is located, $\hat{x}_k^*$ is the complex conjugate of $\hat{x}_k$ which is used to match object's appearance, $\widehat{y_k}$ is a Gaussian-shaped function that represents the expected response where the object is located correctly (i.e., peak where object is located) and $\lambda$ which represents the regularization term.
The generated correlation filter is then applied to all new frames to find the object location through sliding the filter over different regions of the frame and generating the response map. The response map is generated as,

$$r = \mathcal{F}^{-1}\big(\widehat{CF} \cdot \hat{z}\big) \qquad (5)$$

Where $\hat{z}$ is the Fourier transform of the new image region, and the inverse Fourier transformer ($\mathcal{F}^{-1}$) is used to convert the product from frequency domain to spatial domain to locate the object in the new frame.
In [30] an enhancement of the tracker is introduced where it improved through integrating channel and spatial reliability assessment (CSR-DCF). This is achieved by measuring the weights of the important feature channels and spatial regions of the target. Although the enhancement increased the complexity of the system, huge improvements in the robustness of the tracking system is achieved.

## IV. EXPERIMENTAL RESULTS AND DISCUSSIONS

An extensive set of experiments was conducted to demonstrate the effect of simultaneous utilization of all hardware engines available in the Orin GPUs, in addition to integrating a face tracker into the pipeline. An open-source video that consists of 650 frames with a total of 6 celebrity faces [35], in addition to customized and lower-quality videos from the Middle East datasets. The customized video aims to demonstrate the system's ability to handle video streams not taken in ideal situations and to handle traditional headscarves from the Middle East region, as illustrated in Fig.5, which were not considered in earlier systems.

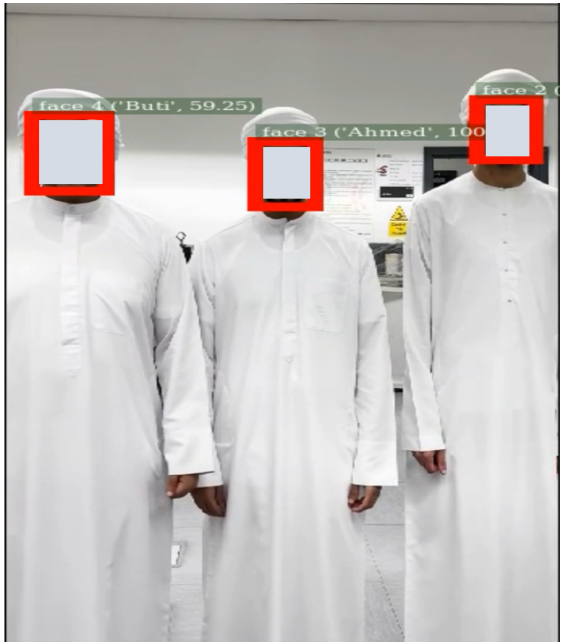

*Figure 5: Frame #743 of a Middle East Test Video*

### A. Hardware Utilization

Four different runs were conducted to study the performance of running the Face Detection and Face Recognition deep learning models on different hardware parts (GPU and DLA). As was mentioned in section III.A, the Jetson AGX Orin comprises two DLAs, namely DLA 0 and DLA 1, where each DLA is assigned to run one of the models. The conducted runs are illustrated in Table.I.

The optimal hardware configuration of the pipeline is selected based on three factors; frame latency, pipeline throughput and GPU/CPU power consumption.

TABLE I
THE FOUR TEST RUNS

| Run No. | Model | Target Hardware | Model | Target Hardware |
|---|---|---|---|---|
| 1 | Face Detection (Face Detect Model) | GPU | Face Recognition (FaceNet Model) | GPU |
| 2 | | DLA 0 | | GPU |
| 3 | | GPU | | DLA 1 |
| 4 | | DLA 0 | | DLA 1 |

#### A.1: Pipeline Throughput

Fig.6 shows the obtained average throughput for the four conducted runs. During each run the average throughput in terms of FPS is captured every 2000ms. All four runs achieved an average throughput that exceeds 30 FPS, proving that the model can achieve real-time performance using the GPU or DLA hardware engines. In addition, from the obtained FPS throughput for the four runs, the effect of parallelism can be observed. To illustrate, when running the full pipeline on the GPU, the average latency for a frame is 75 ms, however the pipeline is able to process 194 frames in one second, this approximately corresponds to 5.2 ms per frame. In similar way, the other 3 runs can be presented by 4.9 ms, 15.8 ms, 16.1 ms, respectively.

The graph shows that the FaceNet model running on DLA decreases the FPS throughput of the pipeline by around 3 times compared to running it on the GPU. This decrease in FPS is anticipated, as explained in Section III.C, mapping FaceNet into DLA results to create 28 shuffle unsupported layers in addition to 28 constant layers and a global average pooling which are not supported. These unsupported layers result to back and forth data transfer between DLA and GPU which introduces transfer latency to each frame processing time. Further, it can as be observed that running the FaceDetect model on DLA slightly increases the throughput by 6 FPS compared to running it on the GPU. Running the FaceDetect on DLA allows parallelism in detection and recognition in the DLA and GPU. Both hardware devices are running simultaneously working on different tasks. Since both devices are using their own memory, no memory conflict is expected during this process. Unlike the FaceNet model, all layers in FaceDetect model are supported by the DLA, and none of the layers fell back to the GPU. Therefore, there is no back and forth data transfer between both devices.

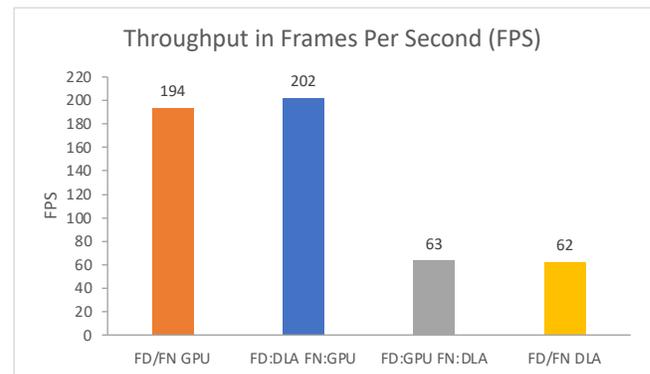

*Figure 6: Effect of Hardware Engine (GPU/DLA) in Throughput (FPS)*

#### A.2: GPU and CPU Power Consumption

Fig.7 shows the power consumption for the GPU and the average power consumption for the 12 CPU cores available in the Jetson Orin during the four different runs. The figure illustrates that running the FaceDetect model on DLA and the FaceNet on the GPU resulted in the lowest power consumption with around 300mW GPU power reduction compared to running both models on the GPU. This is expected as in this case we are offloading the GPU by running all the layers in FaceDetect model on the DLA. Running the FaceNet model on DLA increases further the GPU power consumption because of the unsupported layers of the model which fall back to the GPU. This will result in overhead communication between GPU and DLA that utilizes the GPU power, which increases the average GPU power consumption.

For the CPU power consumption, the CPU average power consumption increases with the activation of more hardware engines in the device. The CPU power consumption increases with the activation of one DLA in the 2nd run, and further increases with the activation of the second DLA as presented in the last run. This increase is because the CPU manages the data transfer process between the hardware parts inside the device and manages the tasks and memory allocations [36]. When the DLA is enabled additional data transfer and memory allocation tasks are introduced. Unlike the first run where data is only transferred from the CPU to the GPU, in the other 3 runs the CPU has to manage data and memory for the two devices; GPU and DLA. The huge increase in the CPU power consumption in the last two runs occur due to the partitioning occurring from falling back of some layers of the model from DLA to GPU.

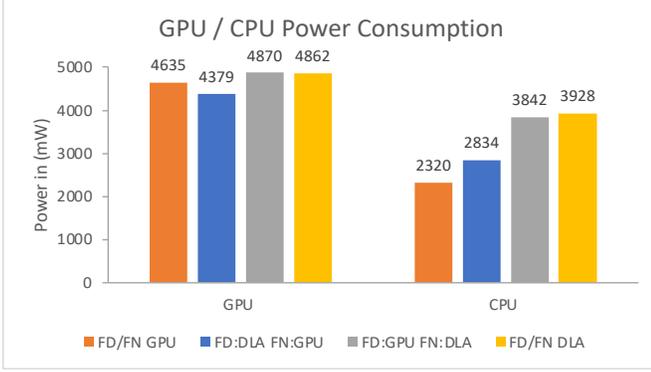

*Figure 7: Effect of Hardware Engine (GPU/DLA) in GPU/CPU Power Consumption*

As illustrated in the aforementioned section, enabling the DLA for face detection model resulted in better throughput performance in addition to a lower power consumption. However, enabling the DLA for the face recognition model introduced latency and higher power consumption. This makes the optimal pipeline being achieved when the face detection model is mapped to the DLA, and the face recognition is running on the GPU

### B. TRACKER INTEGRATION

The three different trackers IOU, SORT and DCF are evaluated in terms GPU/CPU load, GPU/CPU power consumption and robustness. These trackers are evaluated with tracker-by-detection system first, before including the recognition stage to the model.

#### B.1: Face Detection and Tracking
Table.II shows the baseline for comparison where it represents the obtained values from running the face detection pipeline before integrating the tracker. Fig.8 and Fig.9 show the obtained GPU/CPU Load and GPU/CPU power consumption for the three tested trackers, respectively.

TABLE II
THE FOUR TEST RUNS

| Throughput in FPS | GPU Load (%) | CPU Load (%) | GPU Power Consumption (mW) | CPU Power Consumption (mW) |
|---|---|---|---|---|
| 185 | 5.3% | 18.1% | 3506 | 1342 |

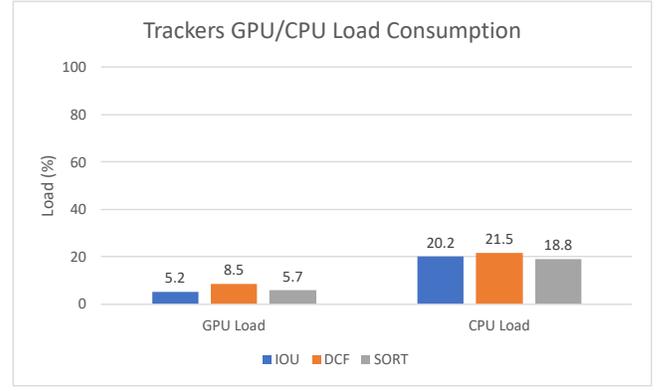

*Figure 8: Trackers GPU/ CPU Load Comparison*

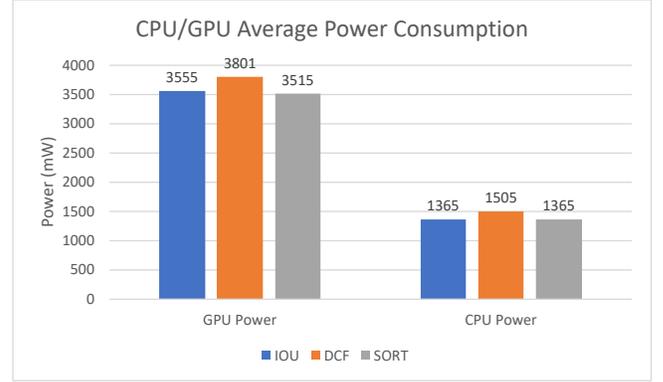

*Figure 9: Trackers GPU/CPU Average Power Consumption*

The graphs clearly illustrate that IOU and SORT trackers almost utilize the same CPU load and CPU power consumption. Since both trackers are light weighted trackers and are based on simple mathematical equations, they can run fully on the VIC without the need for GPU usage. Therefore, these models almost do not introduce any increase in the GPU load or the GPU power consumption. Unlike the IOU and SORT trackers, the DCF uses feature matching which results in robust tracking and less frequent ID switches. However, this performance improvement is at the expense of higher usage of CPU and GPU. Unlike IOU and SORT, the DCF requires low usage of the GPU and cannot run fully on the VIC, this GPU usage is around 3% higher than the other two trackers, with a 300 mW increase in the GPU power consumption. This is because the DCF algorithm involves operations like correlation and Fourier transformers which are more computationally complex compared to intersection simple calculation and Kalman filter. Another test video [37] which includes constant number of 6 faces across all frames is used to analyze tracker performance. A screenshot of one of frames in shown in Fig.10. Starting with the first frame of the video, the faces are first detected and then a unique ID is assigned to each face; from ID=1 for the first detected face in the first frame and ID=6 for the last detected face in the first frame. The tracker will keep track of these detected faces, if the tracker losses the track of any of these faces, a new ID will be assigned to that face, and the face will undergo the recognition stage again. To measure the robustness of the three trackers, the ID switches for faces were monitored for the three trackers. Table.III shows the IOU and SORT trackers performance as both gave the exact same performance, and Table.IV shows a track of the performance of DCF tracker. Whenever the track re-assigns a new ID to any of the faces, this is presented as an "ID switch No." in the table. "Face No." in the table refers to the 6 faces present in the video where Face.1 refers to the first person on the left and Face.6 refers to the person on the right side of the frame as shown in Fig.10.

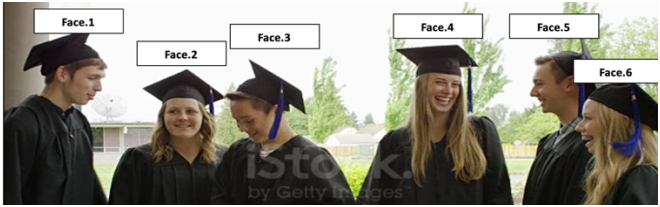

*Figure 10: Tracked faced in a test Video*

In the three trackers, 5 different ID switches took place. For the IOU and SORT algorithms the last ID assigned is 10 which was assigned to Face.5. For faces 1 to 3, IOU and SORT kept an excellent track of these three faces across all frames, that is, if the recognition model is added to the pipeline, faces 1 to 3 will be inputted to recognition stage only once while processing the first frame. For faces 5 and 6, the trackers lost track of the faces 3 times across the video, because face 6 crossed face 5 multiple times which caused the tracker to lose track of both faces. In Table IV, where DCF tracker is used, the last ID assigned is ID=9 to face 5. Face 5 was initially given ID=1 in the first frame, the tracker lost track of the face and then ID=9 was assigned; however, the track was able then to reassign the correct old ID to the same face. The reason behind this, is that, DCF uses visual feature extraction which helps to keep track of the face based on visual and feature similarity [38].

Comparing the three trackers, DCF is more robust and it has less frequent ID switches compared to the other two trackers. However, due to the feature extraction feature it offers it is considered to be more computationally complex and therefore it consumes higher load than the other two trackers. The aim in this paper is to use a robust tracker with no frequent switching. Accordingly, DCF was used for the pipeline in the upcoming section.

*B.2: Face Detection, Tracking and Recognition*

Results in terms of FPS and GPU/GPU power consumptions obtained by this enhancement to the pipeline are presented in Fig.11 and Fig.12, respectively. It is observed that the throughput of the pipeline has increased from 202 FPS to 298 FPS, this increase is due to the reduced redundant recognition processes. This integration of the tracker into the pipeline not only improves the throughput performance but also optimizes the GPU and CPU power consumption. The GPU power consumption is reduced by around 600 mW, where a reduction of around 50% is introduced to the CPU power consumption.

TABLE III
IOU and SORT Trackers ID Switches

| ID Switch No. \ Face No. | Face.1 | Face.2 | Face.3 | Face.4 | Face.5 | Face.6 |
|---|---|---|---|---|---|---|
| 1 | 4 | 3 | 2 | 0 | 1 | 5 |
| 2 | 4 | 3 | 2 | 6 | 7 | 5 |
| 3 | 4 | 3 | 2 | 6 | 7 | 8 |
| 4 | 4 | 3 | 2 | 6 | 8 | 9 |
| 5 | 4 | 3 | 2 | 6 | 10 | 9 |

TABLE IV
DCF Tracker ID Switches

| ID Switch No. \ Face No. | Face.1 | Face.2 | Face.3 | Face.4 | Face.5 | Face.6 |
|---|---|---|---|---|---|---|
| 1 | 4 | 3 | 2 | 0 | 1 | 5 |
| 2 | 4 | 7 | 2 | 6 | 1 | 5 |
| 3 | 4 | 7 | 2 | 6 | 1 | 8 |
| 4 | 4 | 7 | 2 | 6 | 9 | 8 |
| 5 | 4 | 7 | 2 | 6 | 1 | 8 |

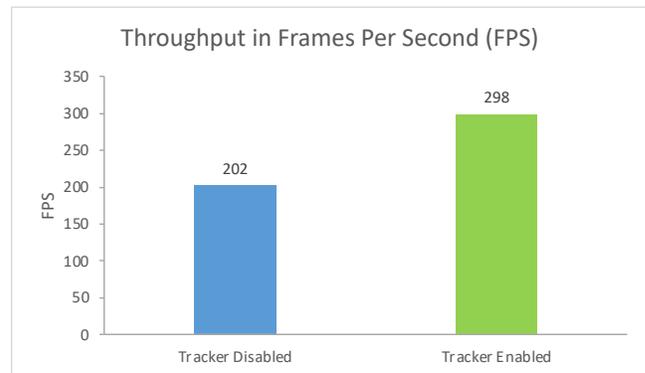

*Figure 11: Tracker Effect on Throughput (FPS)*

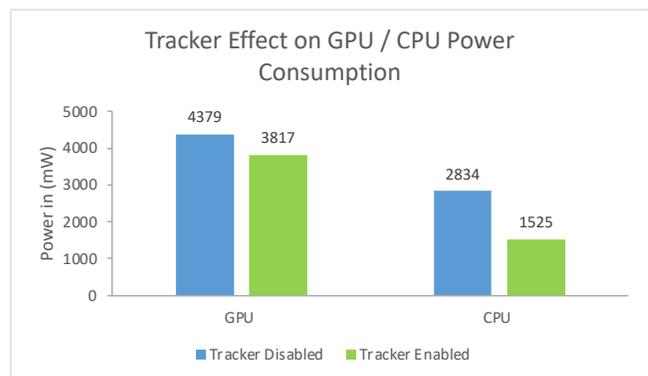

*Figure 12: Tracker Effect on GPU / CPU Power Consumption*

V. DISCUSSION

The developed face detection and face recognition pipeline showed significant improvement to existing studies in terms of throughput and power consumption with no reduction in accuracy. By pruning existing pretrained models to enable them to run on CUDA and Tensor cores, and by utilizing other hardware engines in the device like NVENC and DLA, the pipeline achieved 202 FPS on a test video of 1920x1080 input size with an average of 5 faces per frame. This throughput, which is achieved through hardware utilization alone, before integrating the tracker, is more than 4 times higher than the FPS reported in existing similar studies based on CNN models [20][39][40]. Through hardware optimization the average detection time is around 5.6 ms per frame whereas, average recognition time is around 11.7ms. The average time of detection and recognition is significantly reduced compared to what was achieved in one notable work presented in [20], where different NVIDIA cloud and edge devices were evaluated for a face recognition pipeline, built based on MTCNN for detection and FaceNet for recognition. In [20], Jetson Xavier edge device with 512 CUDA Cores achieved 4 FPS, where RTX 2080 Ti cloud device with 4352 CUDA Cores on the same developed pipeline achieved 20 FPS on three different sizes of the input videos (i.e., 480x480, 1280 x 720, and 1920 x 1080 pixels) with an average of 3.4 faces per frame. As a further improvement to the developed pipeline through hardware utilization, a tracker has integrated between detection and recognition stages. This tracker has allowed to avoid repetitive recognition stage to pre-recognized faces. That is, in many frames where no new faces appear, the recognition stage will have a processing time of 0 ms. Therefore, this reduction in recognition stage time has improved the average FPS from 202 FPS to 300 FPS, and the reduced the total power consumption of CPU and GPU from 8,196 mW to 4,359 mW.

## VI. CONCLUSION

This paper presented an efficient and optimized face detection and recognition pipeline that is designed to leverage the throughput of existing pipelines, considering power. This was achieved through pruning the models to improve the processing time, and by evaluating different hardware configurations and tasks allocation on Jetson AGX Orin edge device. Detailed analysis for each of these configurations was carried out, and it is concluded that the best pipeline is achieved by running the Face Detection model on the DLA, and the Face Recognition model on the GPU. Additionally, a further improvement in the throughput and power consumption was achieved by integrating a tracker between the detection and the recognition stages. This tracker was designed to run on the VIC hardware with very low GPU usage, this avoided repetitive recognition stages and reduced the power consumed by the recognition stage. The tracker improved the throughput of the pipeline from 202 FPS to 298 FPS, in addition to GPU power consumption reduction by 500mW and to more than 1000mW reduction in the CPU power consumption.

## REFERENCES

[1] L. Chen, Y. H. Wang, Y. D. Wang, and D. Huang, "Face recognition with statistical local binary patterns," *Proc. 2009 Int. Conf. Mach. Learn. Cybern.*, vol. 4, no. May 2004, pp. 2433–2439, 2009, doi: 10.1109/ICMLC.2009.5212189.

[2] Kwang In Kim, Keechul Jung, and Hang Joon Kim, "Face recognition using kernel principal component analysis," *IEEE Signal Process. Lett.*, vol. 9, no. 2, pp. 40–42, Feb. 2002, doi: 10.1109/97.991133.

[3] T. Evgeniou and M. Pontil, "Support Vector Machines : Theory and Applications," no. May, 2001, doi: 10.1007/3-540-44673-7.

[4] E. D. D, "Face Recognition using Gabor Filter based Feature Extraction with Anisotropic Diffusion as a pre-processing technique," vol. 45, pp. 312–321, 2015, doi: 10.1016/j.procs.2015.03.149.

[5] S. Bhutekar and A. Manjaramkar, "Parallel face Detection and Recognition on GPU." International Journal of Computer Science and Information Technologies, 2014.

[6] Z. Guo, J. Han, and J. Chen, "Fast face recognition on GPU," in *2015 6th IEEE International Conference on Software Engineering and Service Science (ICSESS)*, 2015, pp. 783–786, doi: 10.1109/ICSESS.2015.7339173.

[7] S. Sapna, R. Anjali, and S. N. Kamath, "Performance Analysis of Parallel Implementation of PCA-based Face Recognition using OpenCL," *2019 4th IEEE Int. Conf. Recent Trends Electron. Information, Commun. Technol. RTEICT 2019 - Proc.*, no. May, pp. 877–881, 2019, doi: 10.1109/RTEICT46194.2019.9016732.

[8] G. Barquero, I. Hupont, and C. F. Tena, "Rank-Based Verification for Long-Term Face Tracking in Crowded Scenes," vol. 3, no. 4, pp. 495–505, 2021, doi: 10.1109/TBIOM.2021.3099568.

[9] M. E. Wibowo, A. Ashari, A. Subiantoro, and W. Wahyono, "Human Face Detection and Tracking Using RetinaFace Network for Surveillance Systems," *IECON 2021 – 47th Annu. Conf. IEEE Ind. Electron. Soc.*, pp. 1–5, doi: 10.1109/IECON48115.2021.9589577.

[10] Z. Weng *et al.*, "Online Multi-Face Tracking With Multi-Modality," *IEEE Trans. Circuits Syst. Video Technol.*, vol. 33, no. 6, pp. 2738–2752, 2023, doi: 10.1109/TCSVT.2022.3224699.

[11] Y. Mao, H. Li, and Z. Yin, "WHO MISSED THE CLASS ? - UNIFYING MULTI-FACE DETECTION , TRACKING AND RECOGNITION IN VIDEOS Department of Computer Science Missouri University of Science and Technology , USA," *2014 IEEE Int. Conf. Multimed. Expo*, pp. 1–6, doi: 10.1109/ICME.2014.6890334.

[12] Z. Zhao, H. Zhang, L. Wang, and H. Huang, "A Multimodel Edge Computing Offloading Framework for Deep-Learning Application Based on Bayesian Optimization," *IEEE Internet Things J.*, vol. 10, no. 20, pp. 18387–18399, 2023, doi: 10.1109/JIOT.2023.3280162.

[13] H. J. Mun and M. H. Lee, "Design for Visitor Authentication Based on Face Recognition Technology Using CCTV," *IEEE Access*, vol. 10, no. October, pp. 124604–124618, 2022, doi: 10.1109/ACCESS.2022.3223374.

[14] S. Manzoor *et al.*, "Edge Deployment Framework of GuardBot for Optimized Face Mask Recognition with Real-Time Inference Using Deep Learning," *IEEE Access*, vol. 10, no. July, pp. 77898–77921, 2022, doi: 10.1109/ACCESS.2022.3190538.

[15] K. Zhang, Z. Zhang, Z. Li, and Y. Qiao, "Joint Face Detection and Alignment Using Multitask Cascaded Convolutional Networks," *IEEE Signal Process. Lett.*, vol. 23, no. 10, pp. 1499–1503, 2016, doi: 10.1109/LSP.2016.2603342.

[16] M. S. Bartlett, J. R. Movellan, and T. J. Sejnowski, "Face recognition by independent component analysis," *IEEE Trans. Neural Networks*, vol. 13, no. 6, pp. 1450–1464, Nov. 2002, doi: 10.1109/TNN.2002.804287.

[17] S. K. BHATTACHARYYA and K. RAHUL, "FACE RECOGNITION BY LINEAR DISCRIMINANT ANALYSIS," *Int. J. Commun. Networks Secur.*, pp. 1–5, Jan. 2014, doi: 10.47893/IJCNS.2014.1087.

[18] F. Schroff, D. Kalenichenko, and J. Philbin, "FaceNet: A unified embedding for face recognition and clustering," *Proc. IEEE Comput. Soc. Conf. Comput. Vis. Pattern Recognit.*, vol. 07-12-June, pp. 815–823, 2015, doi: 10.1109/CVPR.2015.7298682.

[19] A. Baobaid, M. Meribout, V. K. Tiwari, and J. P. Pena, "Hardware Accelerators for Real-Time Face Recognition: A Survey," *IEEE Access*, vol. 10, no. August, pp. 83723–83739, 2022, doi: 10.1109/ACCESS.2022.3194915.

[20] A. Koubaa, A. Ammar, A. Kanhouch, and Y. Alhabashi, "Cloud versus Edge Deployment Strategies of Real-Time Face Recognition Inference," *IEEE Trans. Netw. Sci. Eng.*, vol. X, no. X, 2021, doi: 10.1109/TNSE.2021.3055835.

[21] T. Lindner, D. Wyrwal, M. Bialek, and P. Nowak, "Face recognition system based on a single-board computer," *15th Int. Conf. Mechatron. Syst. Mater. MSM 2020*, pp. 1–6, 2020, doi: 10.1109/MSM49833.2020.9201668.

[22] NVIDIA, "FaceDetect NVIDIA TAO," 2023.

[23] S. Zhang, X. Zhu, Z. Lei, H. Shi, X. Wang, and S. Z. Li, "FaceBoxes: A CPU real-time face detector with high accuracy," in *2017 IEEE International Joint Conference on Biometrics (IJCB)*, 2017, pp. 1–9, doi: 10.1109/BTAS.2017.8272675.

[24] A. Anwar, "Edge-AI based Face Recognition System : Benchmarks and Analysis," *2022 19th Int. Bhurban Conf. Appl. Sci. Technol.*, pp. 302–307, 2022, doi: 10.1109/IBCAST54850.2022.9990546.

[25] H. Aung, B. A. Valentinovich, and B. Aye, "Real-Time Face Tracking Based on the Kalman Filter," in *2022 International Conference on Industrial Engineering, Applications and Manufacturing (ICIEAM)*, 2022, pp. 842–846, doi: 10.1109/ICIEAM54945.2022.9787232.

[26] Q. Zhang, Y. Nie, Y. Wang, and P. Cao, "High-Speed Tracking of Discriminative Correlation Filters Based on GPU," *2023 3rd Int. Conf. Neural Networks, Inf. Commun. Eng. NNICE 2023*, pp. 608–612, 2023, doi: 10.1109/NNICE58320.2023.10105783.

[27] X. F. Zhu, X. J. Wu, T. Xu, Z. H. Feng, and J. Kittler, "Complementary Discriminative Correlation Filters Based on Collaborative Representation for Visual Object Tracking," *IEEE Trans. Circuits Syst. Video Technol.*, vol. 31, no. 2, pp. 557–568, 2021, doi: 10.1109/TCSVT.2020.2979480.

[28] J. Yuan, S. Chen, Z. Shi, and S. Yu, "Discriminative correlation filter tracking algorithm with Transformer based on a multi-frame Cross-Attention mechanism," in *2021 2nd International Conference on Computer Science and Management Technology (ICCSMT)*, 2021, pp. 349–357, doi: 10.1109/ICCSMT54525.2021.00071.

[29] J. F. Henriques, R. Caseiro, P. Martins, and J. Batista, "High-Speed Tracking with Kernelized Correlation Filters."

[30] L. Cehovin, "Discriminative Correlation Filter Tracker with Channel and Spatial Reliability."

[31] T. Xu, Z. H. Feng, X. J. Wu, and J. Kittler, "Learning low-rank and sparse discriminative correlation filters for coarse-to-fine visual object tracking," *IEEE Trans. Circuits Syst. Video Technol.*, vol. 30, no. 10, pp. 3727–3739, 2020, doi: 10.1109/TCSVT.2019.2945068.

[32] NVIDIA, "NVIDIA Deep Learning TensorRT Documentation," 2023. [Online]. Available: https://docs.nvidia.com/deeplearning/tensorrt/quick-start-guide/index.html.

[33] E. Bochinski, V. Eiselein, and T. Sikora, "High-Speed tracking-by-detection without using image information," in *2017 14th IEEE International Conference on Advanced Video and Signal Based Surveillance (AVSS)*, 2017, pp. 1–6, doi: 10.1109/AVSS.2017.8078516.

[34] A. Bewley, Z. Ge, L. Ott, F. Ramos, and B. Upcroft, "Simple online and realtime tracking," in *2016 IEEE International Conference on Image Processing (ICIP)*, 2016, pp. 3464–3468, doi: 10.1109/ICIP.2016.7533003.

[35] "Test Video." [Online]. Available: https://www.youtube.com/watch?app=desktop&si=JIXeAGsgfUiEn0oW&v=UUTn2BzP6mA&feature=youtu.be. Last accessed on 6/1/2024.

[36] "CUDA C++ Programming Guide." [Online]. Available: https://docs.nvidia.com/cuda/cuda-c-programming-guide/index.html.

[37] "Garrett Educational Consulting." [Online]. Available: https://www.garretteducationalconsulting.com/.

[38] NVIDIA, "Gst-nvtracker," 2023. [Online]. Available: https://docs.nvidia.com/metropolis/deepstream/dev-guide/text/DS_plugin_gst-nvtracker.html.

[39] Y. Pan, X. Peng, X. Lin, and M. Xia, "Prototype development of face and speaker recognitions based on edge computing," *2020 Int. Symp. Networks, Comput. Commun. ISNCC 2020*, 2020, doi: 10.1109/ISNCC49221.2020.9297243.

[40] S. Saypadith and S. Aramvith, "Real-Time Multiple Face Recognition using Deep Learning on Embedded GPU System," *2018 Asia-Pacific Signal Inf. Process. Assoc. Annu. Summit Conf. APSIPA ASC 2018 - Proc.*, no. November, pp. 1318–1324, 2019, doi: 10.23919/APSIPA.2018.8659751.